%% file: paper.tex
\documentclass[]{fairmeta}

\definecolor{darkgreen}{rgb}{0.0, 0.5, 0.0}

\newcommand{\ie}

\usepackage{xspace}  % for macros to add space where needed 
\makeatletter
\DeclareRobustCommand\onedot{\futurelet\@let@token\@onedot}
\def\@onedot{\ifx\@let@token.\else.\null\fi\xspace}
\makeatother 
\def\eg{\emph{e.g}\onedot}  
\def\ie{\emph{i.e}\onedot} 
\def\vs{\emph{vs}\onedot}

\newcommand{\repo}{EvalGIM\xspace}

\usepackage{amssymb}
\usepackage{pifont}

\definecolor{saffron}{RGB}{227, 170, 0}
\definecolor{burntOrange}{RGB}{255,122,20}

\definecolor{codegreen}{rgb}{0,0.6,0}
\definecolor{codegray}{rgb}{0.5,0.5,0.5}
\definecolor{codepurple}{rgb}{0.58,0,0.82}
\definecolor{backcolour}{rgb}{0.95,0.95,0.92}

\lstdefinestyle{mystyle}{
    backgroundcolor=\color{backcolour},   
    commentstyle=\color{codegreen},
    keywordstyle=\color{magenta},
    numberstyle=\tiny\color{codegray},
    stringstyle=\color{codepurple},
    basicstyle=\ttfamily\footnotesize,
    breakatwhitespace=false,         
    breaklines=true,                 
    captionpos=b,                    
    keepspaces=true,                 
    numbers=left,                    
    numbersep=5pt,                  
    showspaces=false,                
    showstringspaces=false,
    showtabs=false,                  
    tabsize=2
}

\title{EvalGIM: A Library for Evaluating Generative Image Models}

\author[* 1]{Melissa Hall}
\author[* 1,2]{Oscar Mañas}
\author[* 1]{Reyhane Askari-Hemmat}
\author[* 1]{Mark Ibrahim}
\author[* 1]{Candace Ross}
\author[* 1]{Pietro Astolfi}
\author[* 1,3]{Tariq Berrada Ifriqi}
\author[* 1]{Marton Havasi}
\author[1]{Yohann Benchetrit}
\author[1]{Karen Ullrich}
\author[1]{Carolina Braga}
\author[1]{Abhishek Charnalia}
\author[1]{Maeve Ryan}
\author[1]{Mike Rabbat}
\author[1]{Michal Drozdzal}
\author[1]{Jakob Verbeek}
\author[1,2,4,5]{Adriana Romero-Soriano}

\contribution[*]{Core contributor}

\affiliation[1]{FAIR at Meta}
\affiliation[2]{Mila, Quebec AI Institute}
\affiliation[3]{Univ.\ Grenoble Alpes, Inria, CNRS, Grenoble INP, LJK, France}
\affiliation[4]{McGill University}
\affiliation[5]{Canada CIFAR AI chair}

\abstract{
As the use of text-to-image generative models increases, so does the adoption of automatic benchmarking methods used in their evaluation. 
However, while metrics and datasets abound, there are few unified benchmarking libraries that provide a framework for performing evaluations across many datasets and metrics. 
Furthermore, the rapid introduction of increasingly robust benchmarking methods requires that evaluation libraries remain flexible to new datasets and metrics. 
Finally, there remains a gap in synthesizing evaluations in order to deliver actionable takeaways about model performance. 
To enable unified, flexible, and actionable evaluations, we introduce EvalGIM (pronounced ``EvalGym''), a library for evaluating generative image models.
EvalGIM contains broad support for datasets and metrics used to measure quality, diversity, and consistency of text-to-image generative models.
In addition, EvalGIM is designed with flexibility for user customization as a top priority and contains a structure that allows plug-and-play additions of new datasets and metrics.
To enable actionable evaluation insights, we introduce ``Evaluation Exercises'' that highlight takeaways for specific evaluation questions.
The Evaluation Exercises contain easy-to-use and reproducible implementations of two state-of-the-art evaluation methods of text-to-image generative models: consistency-diversity-realism Pareto Fronts
and disaggregated measurements of performance disparities across groups.
EvalGIM also contains Evaluation Exercises that introduce two new analysis methods for text-to-image generative models: robustness analyses of model rankings and balanced evaluations across different prompt styles.
In this paper, we outline the EvalGIM library and provide guidance for how others can add new datasets, metrics, and visualizations to customize the library for their own use cases. 
We also demonstrate the utility of \repo by using its Evaluation Exercises to explore several research questions about text-to-image generative models, such as the role of re-captioning training data or the relationship between quality and diversity in early training stages.
We encourage text-to-image model exploration with EvalGIM and invite contributions at \url{https://github.com/facebookresearch/EvalGIM/}.
}

\date{\today}
\correspondence{Melissa Hall at \email{melissahall@meta.com}}

\metadata[Code]{\url{https://github.com/facebookresearch/EvalGIM/}}
\metadata[Blogpost]{\url{https://ai.meta.com/blog/meta-fair-updates-agents-robustness-safety-architecture/}}

\begin{document}

\maketitle

\input{sections/introduction.tex}

\input{sections/EvalGIM}

\input{sections/discussion.tex}

\clearpage
\newpage
\bibliographystyle{assets/plainnat}
\bibliography{paper}

\clearpage

\beginappendix
\label{app:appendix}
\input{sections/appendix.tex}
\end{document}

%% file: sections/introduction.tex
\section{Introduction}

The rapid rise of text-to-image generative models has increased the practice of benchmarking their capabilities and weaknesses. 
Even as human evaluations have become an evaluation standard for text-to-image generative models, automatic evaluations remain common, as they allow for timely results about model performance, are easily scaled to many models and datasets, and can be reproduced given a standardized set-up.
For example, it is standard to evaluate models with the  Fréchet Inception Distance (FID)~\citep{heusel2018ganstrainedtimescaleupdate} metric, and increasingly popular to benchmark with precision~\citep{sajjadi2018assessing,kynkaanniemi2019improved} and density~\citep{naeem2020reliable}, which measure fidelity or ``realness" of a set of generated images, and recall~\citep{sajjadi2018assessing,kynkaanniemi2019improved} and coverage~\citep{naeem2020reliable}, which measure the diversity of generated images.
Furthermore, there have been rapid advancements in benchmarks used for measuring image consistency (how well a generated image matches a text prompt), with alignment-based consistency metrics including CLIPScore \citep{hessel2021clipscore}, VQAScore \citep{lin2024evaluating} and VIEScore \citep{ku2023viescore} and question generation-based approaches, such as TIFA \citep{hu2023tifa}, Davidsonian Scene Graph \citep{cho2023davidsonian} and VPEval \citep{cho2024visual}.

\begin{figure}
    \centering
    \includegraphics[width=0.95\linewidth]{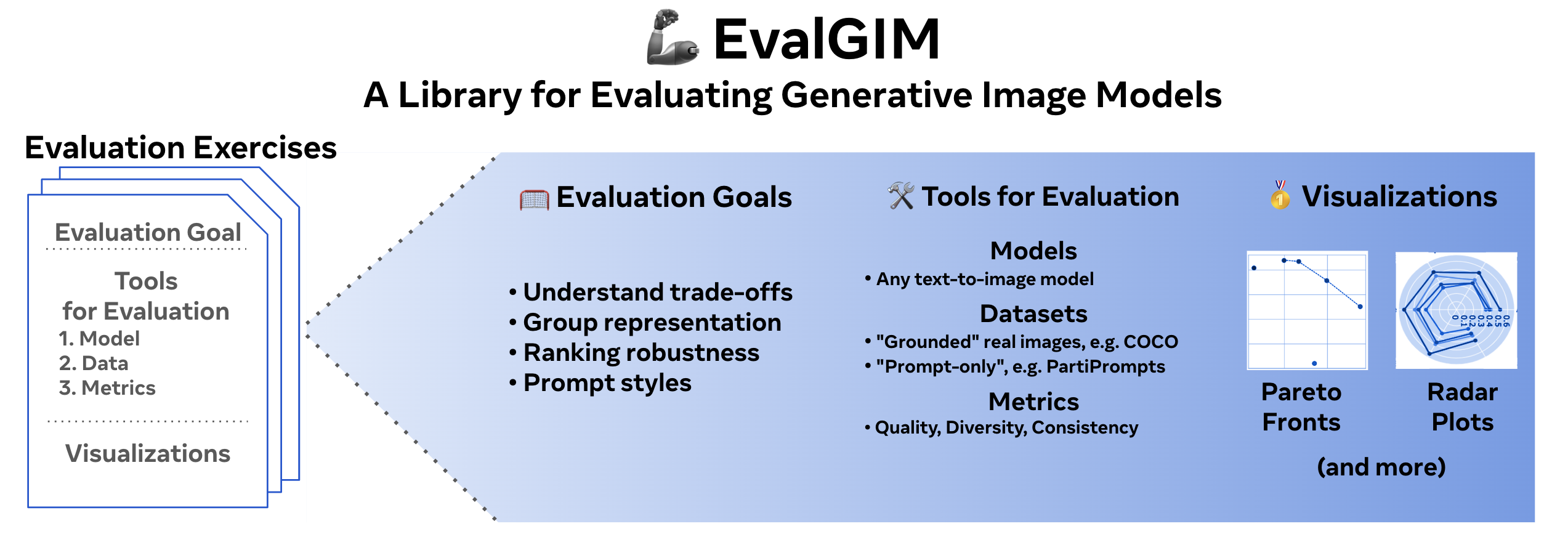}
    \caption{\repo (pronounced as "EvalGym") is an easy-to-use evaluation library for text-to-image generative models that unifies useful evaluation metrics, datasets, and visualizations, supports flexibility for user needs (and extensibility to future benchmarks), and provides actionable insights into model performance. To enable interpretable benchmarking, \repo contains Evaluation Exercises that highlight takeaways for specific evaluation questions related to performance trade-offs, group representation, model ranking robustness, and prompting styles.}
    \label{fig:enter-label}
\end{figure}

While these metrics and benchmarks are useful, their lack of unification makes it inconvenient to perform robust, reproducible evaluations.
In the natural language processing (NLP) domain, it has become quite common to perform multiple evaluations across different model skills with a unified benchmarking library \citep{wang2018glue,wang2019superglue,liang2023holistic}.
However, fewer multifaceted benchmarking libraries have been proposed for the text-to-image domain;
The main benchmarking framework comprising multiple evaluation criteria is HEIM \citep{lee2023holisticevaluationtexttoimagemodels}, which evaluates 12 different aspects, including image-text alignment and reasoning.
Oftentimes, unified libraries present dozens of numbers which, while useful, require extensive study to identify interpretable takeaways about model development.
This can make hypothesis-driven scientific studies challenging. 
Furthermore, they can become outdated quickly, as they rarely introduce newer metrics.
With \repo, we aim to address this gap by providing an easy-to-use evaluation library that unifies evaluation metrics, datasets, and visualizations, supports flexibility for user needs (and extensibility to future benchmarks), and provides actionable insights into model performance.

\repo was designed to be used out-of-the-box, with immediate support for existing models and easy-to-run code targeted for specific evaluation tasks.
To enable this, the library has a unified structure across image generation, evaluation, and visualization, so that resultant data is well-contained and easy to interact with.
Furthermore, \repo easily supports sweeps across different models or hyperparameters and evaluations disaggregated by subgroups of data with the use of a single flag, which provides a more thorough understanding of model performance.
Given the large scale of generative image evaluations, \repo also supports distributed evaluations across compute resources for faster analyses.

In addition, \repo was designed for users to also contribute additional datasets and metrics to expand their analyses. 
We introduce inheritable \texttt{Dataset} types for real image data sources and prompt data sources to enforce consistent image, prompt, and metadata loading.
These classes can be plugged directly into image generation and metric evaluation scripts, allowing for easy additions of new datasets regardless of their source.
Similarly, \repo builds on the \texttt{torchmetrics} library and defines a consistent set of wrapper functions that are used for updating and computing metrics and are compatible with the \texttt{Dataset} types.
This makes it easy to add new metrics to the library such that they work out-of-the box. 

Finally, \repo contains ``Evaluation Exercises'' that allow for structured end-to-end evaluations targetting specific hypotheses about model performance.
These Exercises can be executed in a reproducible manner with user friendly code that requires only a list of model names to execute image generation and evaluation. 
The Exercises return publication-ready visualizations that usefully synthesize takeaways across datasets and metrics and clearly display key results that are relevant to the specific evaluation goals.

In this paper we start by outlining \repo and its customizability for new metrics, datasets, and visualizations. 
Then, we introduce the Evaluation Exercises, which can be executed in a simple three-step process for ease and reproducibility. 
We apply each Evaluation Exercise to a preliminary research analysis, demonstrating how \repo enables progress on text-to-image generative model research. For the first set of Evaluation Exercises, we reproduce evaluation methods of prior works:
\newcommand{\exercise}[1]{\texttt{\textbf{#1}}}
\begin{itemize}
    \item [] \exercise{Evaluation Exercise: Trade-offs.} We study trade-offs in model performance via Pareto Fronts, as introduced by \citet{astolfi2024consistencydiversityrealismparetofrontsconditional}. In our preliminary study with \repo, we find that when training a text-to-image generative model, consistency increases steadily then plateaus, while automatic measures of quality and diversity can fluctuate.
    \item [] \exercise{Evaluation Exercise: Group Representation.} We explore disaggregated representation measurements across geographic groups, as introduced by \citet{hall2024diginevaluatingdisparities}. In our early studies with \repo, we find that advancements in latent diffusion models correspond to an improvement in quality and diversity more for some geographic regions (\eg,  Southeast Asia) than others (\eg,  Africa).
\end{itemize}
Additionally, \repo contains Evaluation Exercises that present new analysis types:
\begin{itemize}
    \item [] \exercise{Evaluation Exercise: Rankings Robustness.} We study the robustness of model rankings across evaluation metrics and datasets. With \repo, we highlight how the relative strengths of candidate models between quality \vs diversity can be obfuscated by the FID metric. In addition, consistency rankings across models can change depending on which metric is used, and model rankings are not always consistent across datasets.
    \item [] \exercise{Evaluation Exercise: Prompt Types.} We perform analyses of balanced comparisons across datasets corresponding to different prompt styles. With \repo, we find that mixing original and re-captioned training data can help improve diversity and consistency.
\end{itemize}

We hope that users find \repo to be easy to use, with flexibility for custom needs and future benchmarks, and helpful in providing actionable insights into model performance. We release \repo to enable future investigations of text-to-image generative models and encourage contributions of new datasets and metrics.

%% file: sections/EvalGIM.tex
\section{EvalGIM for Evaluating Generative Image Models}
\label{sec:library}

Throughout this Section, we describe the different text-to-image generative models, evaluation datasets, metrics, and visualizations supported in \repo.
We also describe how the library can easily be customized by the user to add new evaluation components or state-of-the-art benchmarks.

\subsection{Models and sweeps}
\repo includes support for models that are accessible via the HuggingFace \texttt{diffusers} library. 
This includes functionality to support random seeding across all generations or custom seeding schemes, such as fixed seeds across prompts.
Furthermore, the image generation, evaluation, and visualization scripts include support for hyperparameter sweeps, including over classifier guidance scales, allowing for a deeper understanding of nuances in model performance.

\subsubsection{Add your own models}
While any \texttt{diffusers} model can be run with \repo out-of-the-box, the library can also be integrated directly into existing model training pipelines, allowing for more thorough monitoring of model performance over training time. 
This can be done by creating random latents to sample an existing text-to-image pipeline.
See the repository \texttt{README} for pointers on how to adapt the code for new models. 

\subsection{Evaluation datasets}
\repo contains support for multiple evaluation datasets, including real image datasets and prompt-only datasets. 
We highlight currently supported datasets and how to customize the library by adding more datasets. 
For additional details about dataset construction and filtering, see Appendix \ref{app:datasets}.

\subsubsection{Real image datasets}
Real image datasets have real-world images alongside human-written or automatically generated text prompts. 
\repo includes standard text-to-image evaluation datasets \textbf{MS-COCO}~\citep{lin2014microsoft}, \textbf{ImageNet}~\citep{5206848}, and \textbf{Conceptual 12M (CC12M)}~\citep{changpinyo2021cc12m}.
Additionally, to provide insights into performance for different geographic regions, \repo includes support for the \textbf{GeoDE}~\citep{ramaswamy2023geodegeographicallydiverseevaluation} dataset.
GeoDE contains images of objects that were taken by people living in different countries around the world and has been used to evaluate text-to-image generative models for performance gaps across geographic regions~\citep{hall2024diginevaluatingdisparities,sureddy2024decomposedevaluationsgeographicdisparities}.

\subsubsection{Prompt datasets}
Prompt datasets contain only text used for conditioning image generation.
\repo supports the \textbf{PartiPrompts}~\citep{yu2022scalingautoregressivemodelscontentrich} dataset containing $1600$ prompts of varying complexity and theme.
It also supports compositionality benchmarks \textbf{T2I-Compbench} ~\citep{huang2023t2icompbenchcomprehensivebenchmarkopenworld} and \textbf{DrawBench}~\citep{saharia2022photorealistictexttoimagediffusionmodels}.

\subsubsection{Add your own datasets}
\repo supports the addition of custom datasets. 
For additional real image datasets, developers can leverage the \texttt{RealImageDataset()} class, which requires a set of real images and optionally supports class- or group-level metadata. 
To incorporate new prompts used for image generation, developers may use the \texttt{RealAttributeDataset()} class, which requires prompt strings.
This class also optionally supports class- or group-level labels and metadata corresponding to metric calculation, such as question-answer graphs.

\begin{lstlisting}[language=Python,caption=New image and prompt datasets can easily be added to \repo.]
    
class RealImageDataset(ABC):
    """Dataset of real images, used for computing marginal metrics.
    """

    @abstractmethod
    def __getitem__(self, idx) -> RealImageDatapoint:
        """Returns RealImageDatapoint containing
            image: Tensor
            class_label: Optional[str]
            group: Optional[List[str]]
        """

class RealAttributeDataset(ABC):
    """Dataset of prompts and metadata, used for generating images and computing metrics.
    """

    @abstractmethod
    def __getitem__(self, idx) -> RealAttributeDatapoint:
        """Returns RealAttributeDatapoint containing
            prompt: str
            condition: Condition
            class_label: Optional[str]
            group: Optional[List[str]]
            dsg_questions: Optional[List[str]]
            dsg_children: Optional[List[str]]
            dsg_parents: Condition | None = None
        """
\end{lstlisting}

\subsection{Metrics}

\repo provides support for many evaluation metrics. We highlight marginal metrics, conditional metrics, and grouped metrics included in \repo and discuss how users can add their own metrics. 
We provide additional details about the metrics in Appendix \ref{app:metrics}.

\subsubsection{Marginal metrics: Image Realism \& Diversity}
Marginal metrics measure model performance by comparing distributions of images that are generated with prompts corresponding to real world images to the distribution of real images.
We include \textbf{Fréchet Inception Distance} (FID)~\citep{heusel2018ganstrainedtimescaleupdate}, a standard evaluation metric that compares a distribution of generated images to a corresponding set of real images to provide an indicator of generative quality and diversity. 
To support more detailed insights into trade-offs in model performance, we include \textbf{precision}~\citep{Sajjadi2018_PR} and \textbf{density}~\citep{kynkäänniemi2019improved} as indicators of generated image realism and \textbf{recall}~\citep{Sajjadi2018_PR} and \textbf{coverage}~\citep{kynkäänniemi2019improved} as signals of generated image diversity.

\subsubsection{Conditional metrics: Image-Text Consistency}
Conditional metrics evaluate generated images while leveraging the prompt used in its conditioning, without any grounding with real-world images. 
\repo includes consistency metrics that measure how well generated images match the text prompt used in its generation.
\textbf{CLIPScore}~\citep{hessel2021clipscore} embeds the generated image and text prompt and uses a CLIP model~\citep{Radford2021LearningTV} to measure the cosine similarity between the two embeddings.
\textbf{Davidsonian Scene Graph} (DSG)~\citep{cho2023davidsonian} leverages a language model to generate questions based on the text prompt and a visual question answering (VQA) model to answer the generated questions given the generated images. 
\repo also includes the \textbf{VQAScore}~\citep{lin2024evaluating}, which uses a VQA model to predict the alignment between a text prompt and a generated image.

\subsubsection{Grouped metrics}
 \repo builds on prior works that study disparate performance between groups ~\citep{hall2024diginevaluatingdisparities,sureddy2024decomposedevaluationsgeographicdisparities} and automatically supports disaggregated group-level measurements.
These evaluations are supported for all prompt datasets that contain group information, including subpopulations (\ie geographic terms used when prompting), demographic groups (\ie gender-presentation based on prompting that is used), and dataset partitions (\ie color- or shape-specific prompts).

\subsubsection{Add your own metrics}
\repo builds on \texttt{torchmetrics}~\citep{detlefsen2022torchmetrics} as the framework for distributed metric calculations. 
Intermediate metric values are updated with each batch of real or generated images then computed holistically with the \texttt{compute()} function once all batches have been added. 
For marginal metrics, we introduce two primary functions for updating metric values at each batch: \texttt{update\_real\_images()} and  \texttt{update\_generated\_images()}  which take as inputs batches of real and generated images, respectively, and their associated metadata. 
Conditional metrics use only an \texttt{update()} function, which is equivalent to \texttt{update\_generated\_images()}. 

\begin{lstlisting}[language=Python,caption=New metrics can easily be added to \repo by leveraging the \texttt{torchmetrics} library.]    
class MarginalMetric(Metric):
    def update_real_images(
        self,
        reference_images: torch.Tensor,
        real_image_datapoint_batch: dict,
    ) -> None:
    
    def update_generated_images(
        self,
        generated_images: torch.Tensor, 
        real_attribute_datapoint_batch: dict
    ) -> None:

    def compute(self) -> dict:
        return {"metric_name": super().compute()}

class ConditionalMetric(Metric):
    def update(
    		self, 
        generated_images_batch: torch.Tensor, 
        real_attribute_datapoint_batch: dict
    ) -> None:

    def compute(self) -> dict:
        return {"metric_name": super().compute()}
\end{lstlisting}

Additionally, the \repo supports the addition of metrics that can be disaggregated by subgroups.
For information on how to adapt new metrics to support grouped measurements, see the \texttt{README} of \repo.

\subsection{Visualizations}
The \repo contains scripts for creating reproducible, easy-to-read visualizations.

\subsubsection{Pareto Fronts} 
We implement Pareto Fronts visualizations, which were introduced in the context of understanding trade-offs in text-to-image model performance in~\citet{astolfi2024consistencydiversityrealismparetofrontsconditional}.
These visualizations demonstrate the relationship between improvements of different metrics and can be used to easily plot many datapoints across model types or image generation hyperparameters. 
An example Pareto Front is shown in Figure \ref{fig:tradeoffs}.

\subsubsection{Radar plots}
While Radar plots have been used for text-to-image model performance across different metrics~\citep{flux}, we extend their utility by leveraging them for group-level measurements.
In \repo, the radial axes correspond to group performance, and the relative location of groups along their axes allows for comparison of disparities in performance across groups. 
Furthermore, multiple models can be included on a plot to study which groups realize the most improvement. 
An example Radar plot is presented in Figure \ref{fig:representation}.

\subsubsection{Ranking table}
\begin{figure}[ht!]
    \centering
    \includegraphics[width=0.98\linewidth]{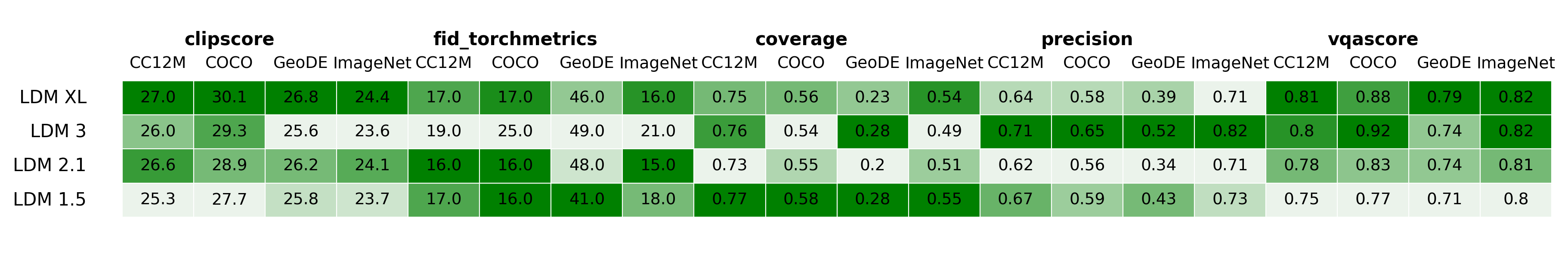}
    \caption{\textbf{Utilizing the Ranking Robustness Evaluation Exercise across a suite of metrics and datasets reveals information that may be obfuscated by individual metrics.}
    When applied to preliminary studies of latent diffusion models, we find that which models have better quality \textit{v.s.} diversity can be obfuscated by FID. 
    In addition, which model has the worst consistency can change depending on which consistency metric is used, and model rankings are not always consistent across datasets.}
    \label{fig:ranking}
\end{figure}

We build on previous works \citep{lee2023holisticevaluationtexttoimagemodels} that present large tables of results across metrics and models by introducing ranking table visualizations. 
We also include multiple datasets for the same metric to provide insights into whether model rankings are consistent for different data distributions.
\repo provides additional interpretability by applying color coded indicators of model rankings across a given dataset and metric, visually revealing the robustness of a model's ranking.
An example is shown in Figure \ref{fig:ranking}.

\subsubsection{Scatter plots} 
\begin{figure}[ht!]
  \centering
  \begin{minipage}[b]{0.33\textwidth}
    \includegraphics[width=\textwidth]{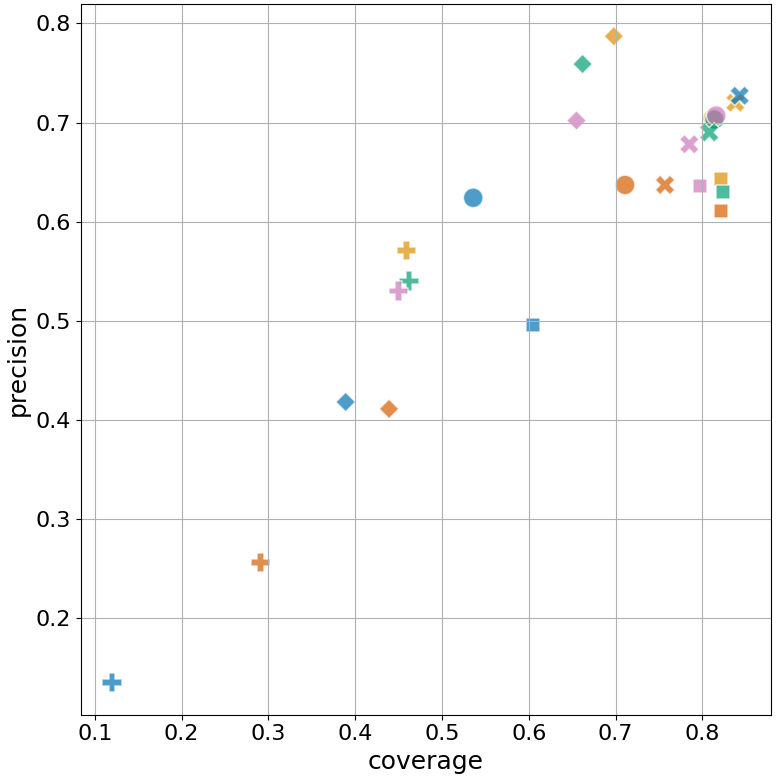}
  \end{minipage}
  \hfill
  \begin{minipage}[b]{0.66\textwidth}
    \includegraphics[width=\textwidth]{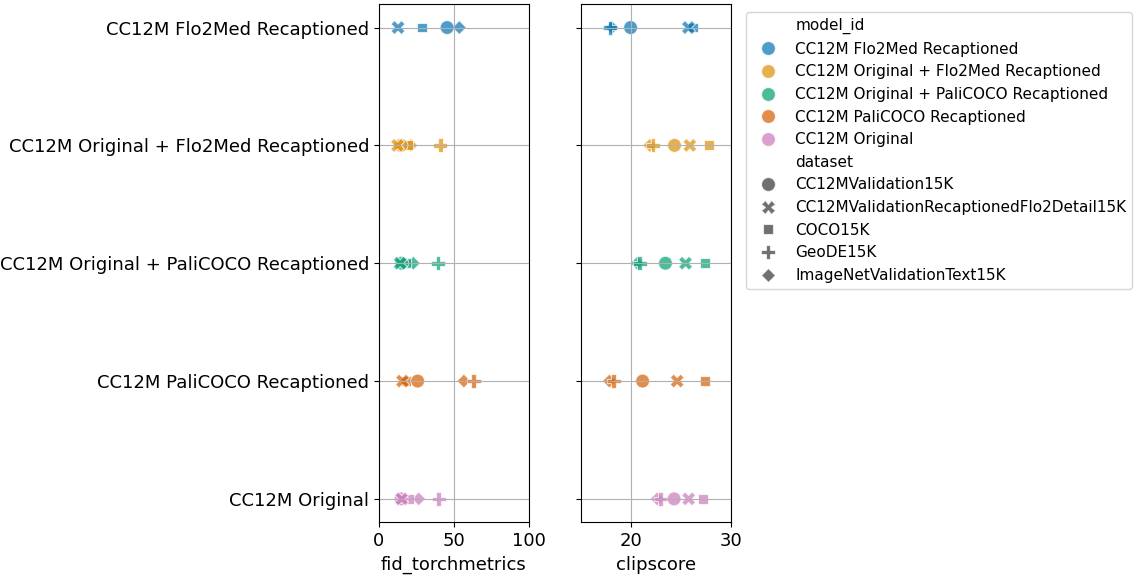}
  \end{minipage}
  \caption{\textbf{Analyzing models with the Prompt Types Evaluation Exercise provides insights into how training time interventions affect certain datasets more than others.}
  When applied to a study of training text-to-image models with recaptioned training data, we find that mixing original and recaptioned training data can help improve diversity and consistency. 
  }
  \label{fig:scatter}
\end{figure}

We also include scatter plots to compare performance across datasets that are subsampled to be the same size, enabling a fair comparison of marginal metrics between them.
An example is presented in Figure \ref{fig:scatter}.

\subsubsection{Add your own visualizations}
Visualization scripts are stored in the \texttt{visualizations/} directory of \repo and take as inputs a \texttt{csv} with evaluation results for each model-dataset-hyperparameter combination and possible visualization parameters, such as a list of metrics to display. 
New visualizations can be added following this scheme.

\section{Analyzing Models with Evaluation Exercises}
\label{sec:exercises}
There are hundreds of different combinations of text-to-image model evaluations possible across dataset, metric, and visualization options in \repo. 
A standard practice would be to either focus on a single dataset-metric-visualization combination for ease of actionability, at the cost of understanding the full picture of model performance or generate a large table of metric results across different axes of evaluations, to the detriment of interpretatability and actionability. 

To address this gap, \repo contains ``Evaluation Exercises'' for common analysis questions about text-to-image generative models.
For example, a common evaluation question is, ``What is the trade-off between quality, diversity, and consistency?'' for a given model. 
Each Evaluation Exercise organizes a specific set of datasets, metrics, and visualizations to provide a clean, easy-to-interpret insight about text-to-image performance. 
While each Exercise could be extended with alternative datasets and metrics, we propose a specific subset here which yield reliable and interpretable analyses. 
To help with ease-of-use, each Evaluation Exercise runs with a simple notebook to generate images, evaluate, and visualize the main takeaways. 

\begin{figure}[ht]
    \centering
    \includegraphics[width=0.85\linewidth]{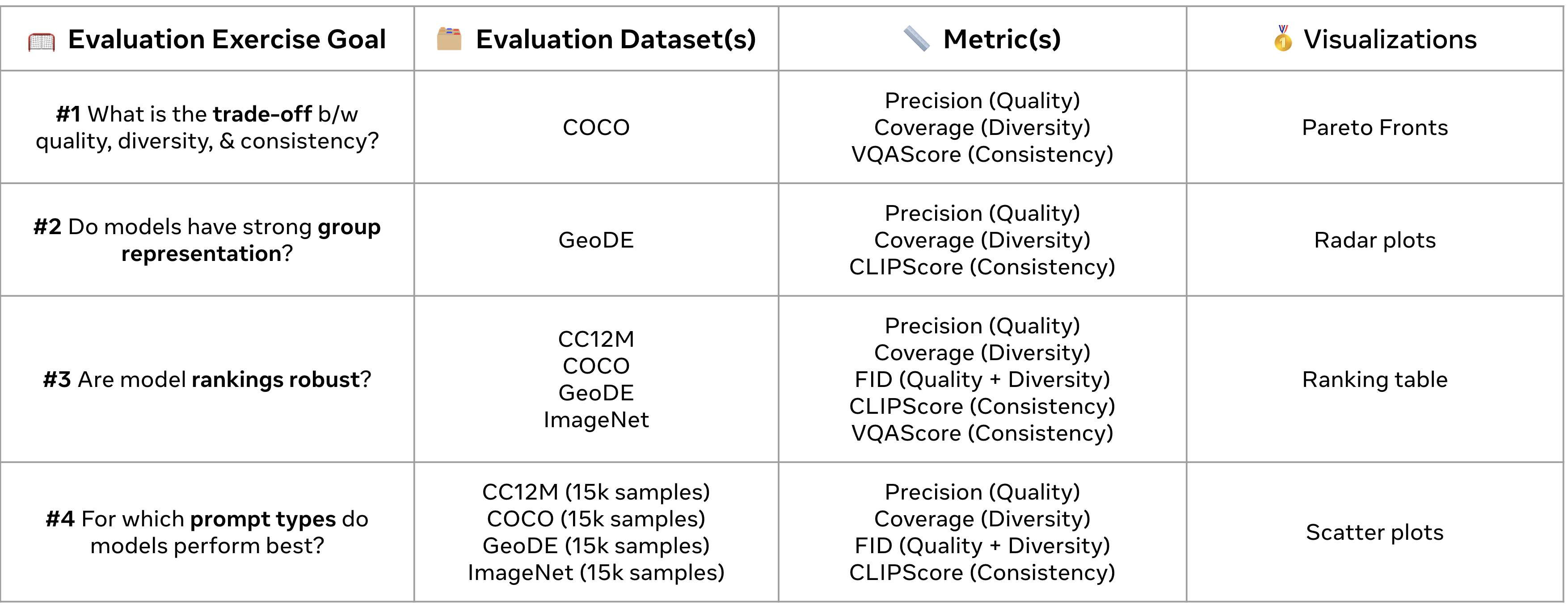}
    \caption{\repo contains Evaluation Exercises that allow for structured end-to-end evaluations targeting specific analysis goals. The Exercises can be easily executed in a reproducible manner with user friendly notebooks.}
    \label{fig:exercises}
\end{figure}

The Evaluation Exercises are summarized in Figure \ref{fig:exercises}, and we describe them in further detail throughout this section. 
Additionally, we demonstrate their ease-of-use and interpretability by applying each one to a preliminary research analysis.

\subsection{Evaluation Exercise: Trade-offs}

This Exercise explores tradeoffs across quality, diversity, and consistency of generated images.
It is intended to validate and make reproducible analyses that have been previously explored by~\citet{astolfi2024consistencydiversityrealismparetofrontsconditional}.

\subsubsection{Exercise definition}
\paragraph{Datasets.}
Because this Evaluation Exercise leverages marginal metrics, it requires a dataset that contains real images. 
We use COCO, which is a standard in text-to-image generation evaluations and contains human-written, realistic captions that can be used as prompts. 

\paragraph{Metrics.}
Alternatives to FID are used in order to differentiate between improvements in quality, diversity, and consistency. 
Precision is used to measure generated image quality, as it quantifies the proportion of generated images that fall within a manifold of real images. 
Coverage is used to quantify how well generated images represent the diversity of real images. 
To measure consistency, we depart from prior works and use VQAScore, which contains a higher correlation with human preference of image consistency compared to alternative metrics like CLIPScore~\citep{lin2024evaluating}. 

\paragraph{Visualizations.}
Following ~\citet{astolfi2024consistencydiversityrealismparetofrontsconditional}, we leverage Pareto Fronts drawn across different consistency, diversity, and quality metrics.
These Pareto Fronts visualize trade-offs across different interventions, such as ablations (\eg model size), hyperparameter sweeps (\eg guidance scale), or training progress (\eg iterations). 

\subsubsection{Exercise in action}
As an example of the utility of the Trade-offs Evaluation Exercise, we use it to explore how a model varies in quality, diversity, and consistency throughout its early training process.

\begin{figure}[th!]
    \centering
    \includegraphics[width=0.9\linewidth]{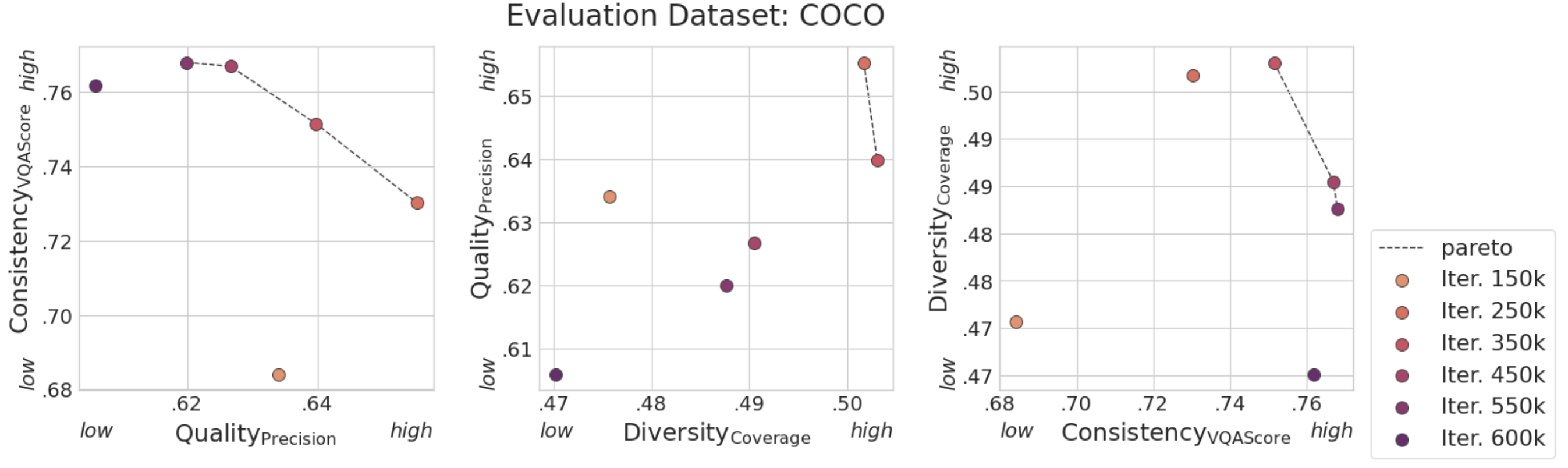}
    \caption{\textbf{Utilizing the Trade-offs Evaluation Exercise gives insights into the relationship between quality, diversity, and consistency.} 
    When applied to preliminary studies of early training of a text-to-image generative model, consistency (as measured by VQAScore) increases then plateaus, while automatic measures of quality and diversity can fluctuate.}
    \label{fig:tradeoffs}
\end{figure}

\paragraph{Experimental Set-up}
We train a text-to-image generative model that leverages flow matching ~\citep{lipman2023flowmatchinggenerativemodeling} and control conditions~\citep{podell2023sdxlimprovinglatentdiffusion,ifriqi2024improvedconditioningmechanismspretraining} with a dataset of image-caption pairs including a subset from ImageNet~\citep{5206848}, CC12M~\citep{changpinyo2021cc12m}, and an internally licensed dataset.
The model is trained for a total of $650,000$ iterations and evaluated with COCO using a classifier guidance scale of $7.5$.
Results are shown in Figure \ref{fig:tradeoffs}.

\begin{figure}[!ht]
  \centering
    \begin{minipage}[b]{0.49\textwidth}
    \includegraphics[width=\textwidth]{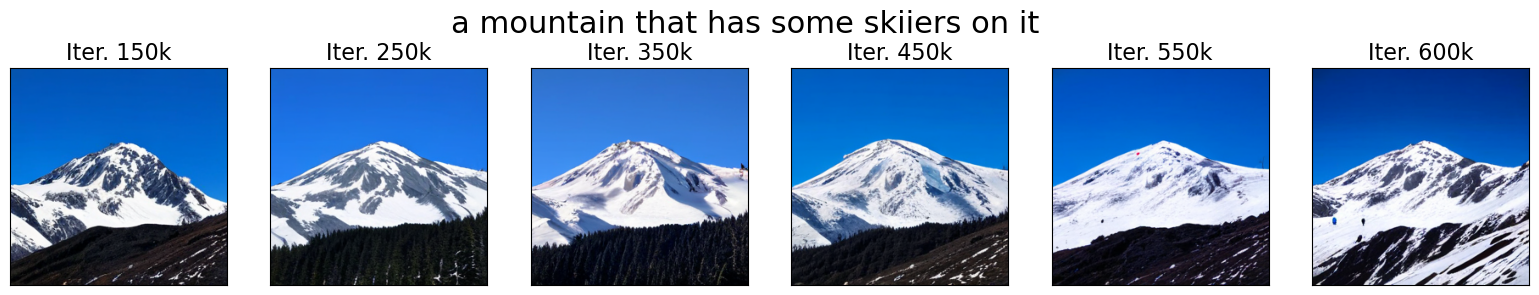}
  \end{minipage}
  \hfill
  \begin{minipage}[b]{0.49\textwidth}
    \includegraphics[width=\textwidth]{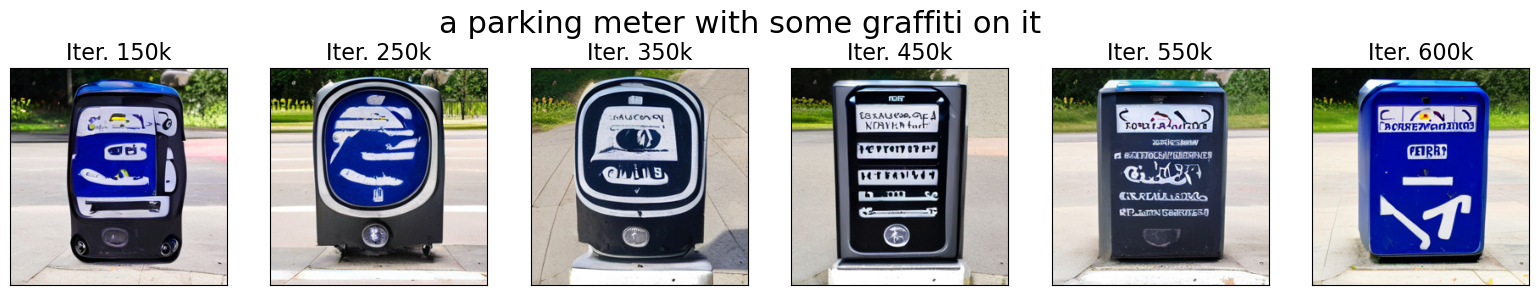}
  \end{minipage}
    \begin{minipage}[b]{0.49\textwidth}
    \includegraphics[width=\textwidth]{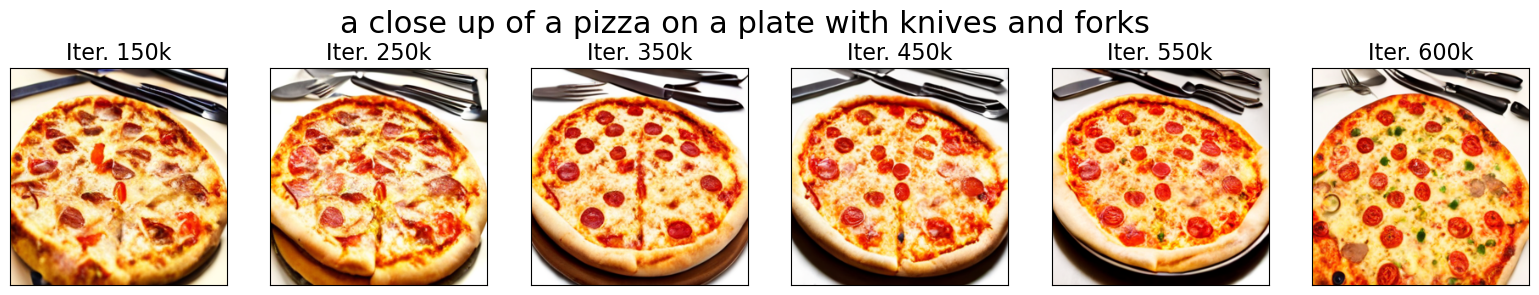}
  \end{minipage}
  \hfill
  \begin{minipage}[b]{0.49\textwidth}
    \includegraphics[width=\textwidth]{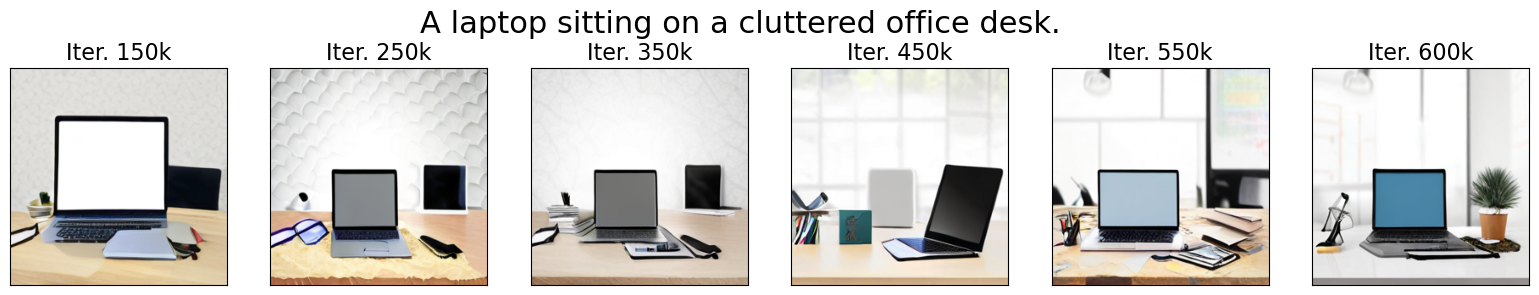}
  \end{minipage}

  \caption{Finer-detailed image improvements identified via qualitative inspection do not always translate to improvements in automatic measures of quality and diversity.}
  \label{fig:tradeoffs_visual}
\end{figure}

\paragraph{Findings}
We find that throughout the training process, quality, diversity, and consistency trade-off in different ways and are not always zero-sum. 
Consistency steadily increases initially in the training process, then plateaus at around $450,000$ iterations.
Quality (as measured by precision) can have a small but steady decrease, perhaps as generation capabilities increase beyond the distribution of the reference dataset used for evaluation. 
Furthermore, representation diversity shows an initial small increase then a slight decrease.
Improvements in consistency coincide with small declines in quality.
In addition, we inspect visual samples (see random examples in Figure \ref{fig:tradeoffs_visual}) across the models and find that image-prompt alignment often visibly improves over the training process, aligning well with improvements in the consistency metric.
Throughout training the images show improvements in finer-grained details, realistic shapes, and well-formed structures. 
It is left to future work to explore how these trends evolve over future iterations.

\subsection{Evaluation Exercise: Group Representation}
\label{sec:group}

This Evaluation Exercise studies performance of models for different subgroups of data, leveraging the disaggregated evaluation method presented by~\citet{hall2024diginevaluatingdisparities}.

\subsubsection{Exercise definition}

\paragraph{Datasets}
We use the GeoDE dataset, disaggregated at the regional level. 
This corresponds to the geographic representation of different objects like ``car'' and ``cooking pot'' across regions around the world.

\paragraph{Metrics}
Precision and coverage are used to measure quality and diversity and CLIPScore for a measure of consistency.
While prior work~\citep{hall2024diginevaluatingdisparities} uses the bottom 10th-percentile CLIPScore, we focus on CLIPScore for the full prompt.
We note that CLIPScore may give higher ratings to more stereotypical representations, as suggested in previous work~\citep{agarwal2021evaluatingclipcharacterizationbroader,10.1145/3630106.3658927}.

\paragraph{Visualization}
Radar plots are used for each metric, with axes corresponding to group performance. 
The relative location of groups along their axes allows for comparison of disparities in performance across group.
Multiple models can be included on a plot to study which groups experience the most improvement, as indicated by the gap on a given group's axis between model versions.

\subsubsection{Exercise in action}

Using the Group Representation Evaluation Exercise, we explore the evolution of geographic representation across different generations of the same family of latent diffusion models.

\paragraph{Experimental Set-up}

We evaluate different generations of a latent diffusion model (\texttt{LDM}).
The first, \texttt{LDM-1.5}~\citep{Rombach_2022_CVPR}, is trained on a public web dataset containing approximately 2 billion images and further trained on higher resolution images and fine-tuned on aesthetic images. 
The second, \texttt{LDM-2.1}~\citep{Rombach_2022_CVPR}, is similarly trained on a dataset of approximately 5 billion images and further trained for multiple iterations on increasingly high-resolution samples, then fine-tuned.
We also evaluate \texttt{LDM-3}~\citep{esser2024scalingrectifiedflowtransformers}, which incorporates rectified flows~\citep{liu2022flow,albergo2022building,lipman2023flowmatchinggenerativemodeling} and leverages three pre-trained text encoders. 
Finally, we evaluate \texttt{LDM-XL}~\citep{podell2023sdxlimprovinglatentdiffusion}, a base model to that generates latents using a larger UNet backbone and more parameters than the aforementioned models. 
We access all models via an API containing open-sourced model weights.

\paragraph{Findings}

\begin{figure}[th!]
    \centering
    \includegraphics[width=0.9\linewidth]{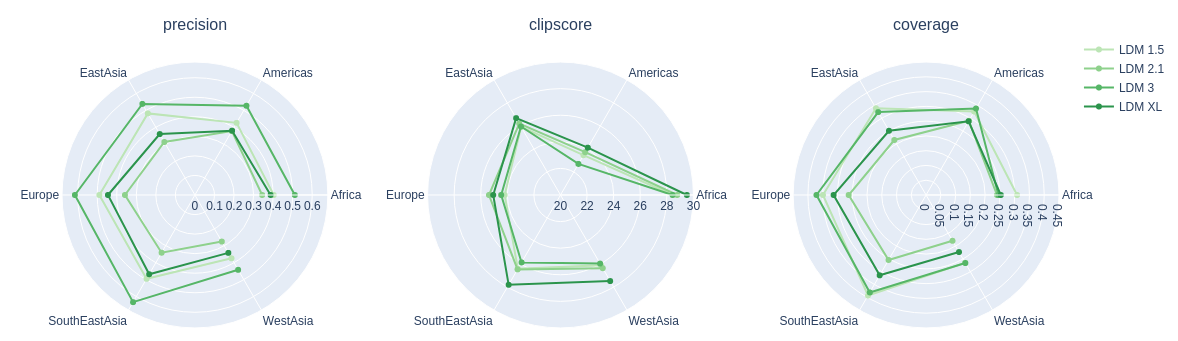}
    \caption{\textbf{Using the Group Representation Evaluation Exercise provides insights into potential disparities in model performance across groups and whether improvements over successive model generations have occurred similarly across groups.}
    When studying successive versions of a latent diffusion model with increasingly complex training data and fine-tuning methods, we find that advancements correspond to an improvement in quality and diversity more for some geographic regions (\eg Southeast Asia) than others (\eg Africa).}
    \label{fig:representation}
\end{figure}

With the Group Representation Exercise, we can study the gaps along each region's radial axis to understand how modeling advances affect groups disparately. 
Figure \ref{fig:representation} shows that there has been more improvement in the quality (precision) of images from Southeast Asia  and Europe compared with those in West Asia and Africa.
Notably, while previous work on earlier version of latent diffusion models highlighted \textit{drops} in diversity and quality when studying geographic representation over successive versions of models~\citep{hall2024diginevaluatingdisparities,astolfi2024consistencydiversityrealismparetofrontsconditional}, we find the most recent model, \texttt{LDM-3}, \textit{recovers} most of the performance loss of previous models.
The one exception is for representational diversity for Africa, where \texttt{LDM-3} lags behind the older \texttt{LDM-1.5} in coverage measurements.

\subsection{Evaluation Exercise: Ranking Robustness}
\label{sec:analysis_ranking}
We also introduce new evaluation formulations to improve understanding of text-to-image models. 
This Ranking Robustness Evaluation Exercise studies whether relative performance across different candidate models is consistent across a set of metrics and datasets.
It is intended to serve as a more interpretable variant of large metric tables that are often provided in benchmarking efforts and proposals of new models, with a focus on highlighting whether claims of superior performance for a given model are actually consistent across different metrics and datasets. 

\subsubsection{Exercise definition}

\paragraph{Datasets}
We recommend both well-established datasets and more recent variants that have been demonstrated to provide contextualization of standard benchmarks. 
For established datasets we include ImageNet and COCO. 
For newer datasets, we include GeoDE, which helps highlight whether overall improvements come at the cost of geographic representation, and CC12M, which uses alt-text captions as image prompts.

\paragraph{Metrics}
We recommend a combination of metrics that are included in other libraries, such as FID and CLIPScore~\citep{lee2023holisticevaluationtexttoimagemodels}, and additional, newer variants such as precision, coverage, and VQAScore to better contextualize performance claims. 
We include FID and additionally supplement with precision and coverage to demonstrate whether advances in quality or diversity, respectively, contribute to improvements in FID. 
For consistency, we include CLIPScore, which is more established, and the newer VQAScore, which has a stronger correlation with human annotations~\citep{lin2024evaluating}.

\paragraph{Visualizations}
This Evaluation Exercise is visualized with a ranking table across all datasets and metrics, where each cell is colored according to how the model ranks for the respective metric-dataset combination. 
Thus, it allows for easier identification of which models perform best across specific metrics or datasets and whether rankings across models are consistent.

\subsubsection{Exercise in action}

We study whether rankings of latent diffusion models are robust across different metrics and datasets.

\paragraph{Experimental set-up}

We study the same set of latent diffusion models described in Section \ref{sec:group}: \texttt{LDM-1.5}, \texttt{LDM-2.1}, \texttt{LDM-XL}, and \texttt{LDM-3}.

\paragraph{Findings}

The Rankings Robustness Evaluation Exercise provides insight into how information about model performance may be obfuscated across different metrics. 
We find that disaggregating measurements between quality (\ie precision) and diversity (\ie coverage) can tell a different story than combined measurements (\ie FID).
For example, we observe that \texttt{LDM-3} ranks worst on FID across all datasets but highest for quality. 
Similarly, \texttt{LDM-1.5} scores quite well on FID but this is likely due to its stronger diversity rather than quality. 
Furthermore, different consistency metrics can yield different outcomes of ``best'' models: when evaluating with GeoDE and ImageNet,  \texttt{LDM-1.5} shows the worst consistency when evaluated with VQAScore while \texttt{LDM-3} is worst when using CLIPScore (although the two are quite close).
In addition, this analysis provides insights into the robustness of metrics across different datasets.
For example, quality (\ie precision) and consistency (\ie CLIPScore and VQAScore) rankings correlate well across datasets.
However, diversity (coverage) rankings for the models vary when evaluating with different datasets.
For example, \texttt{LDM-3} has among the strongest diversity when evaluated on the geographically representative GeoDE dataset but lowest on COCO and ImageNet.

\subsection{Evaluation Exercise: Prompt Types}
\label{sec:prompt_types}
The Prompt Types Evaluation Exercise focuses on model performance across different prompt types, from a single concept word such as ``apple'' or ``dog'' to multi-sentence descriptions that capture details such as image styles, multiple objects, and layout.
Analyzing text-to-image models along these different prompt types provides insights into whether a given intervention helps certain kinds of model interaction types (via prompting) moreso than others.

\subsubsection{Exercise definition}

\paragraph{Datasets}
In a departure from prior works, we balance evaluation datasets so that they can be compared to each other with distributional metrics. 
In our case, we uniformly randomly subsample ImageNet, GeoDE, COCO, and CC12M so that they are the largest possible shared sized, e.g. approximately $15,000$ images (as dictated by the size of the smallest dataset, CC12M). 
ImageNet prompts correspond to individual concepts, GeoDE prompts outline objects in a given geographic region, and COCO prompts are human-written captions of the corresponding real image. 
For CC12M, we use two types of prompts: the original alt-text based captions for the corresponding image and a re-captioned version using the outputs of the Florence-2 model~\citep{xiao2023florence2advancingunifiedrepresentation} when provided with the respective real image.

\paragraph{Metrics}
We use precision and coverage to measure quality and diversity and CLIPScore for consistency.

\paragraph{Visualization}
We include a scatterplot of precision \textit{vs.} coverage across the evaluation datasets, which are comparable due to being balanced.
We also include scatterplots of FID and CLIPScore.

\subsubsection{Exercise in action}

We use the Prompt Types Evaluation Exercise to perform an initial study of how different training data re-captioning methods can affect the diversity of generated images. 

\paragraph{Experimental Set-up}
We focus on a latent diffusion model that has been trained with image-caption pairs from CC12M~\citep{changpinyo2021cc12m} while employing different re-captioning strategies: 
Images are recaptioned with either PaliGemma ~\citep{beyer2024paligemmaversatile3bvlm} (fine-tuned on COCO) or Florence-2 ~\citep{xiao2023florence2advancingunifiedrepresentation}. 

\paragraph{Findings}

\begin{figure}[ht!]
  \centering
  \begin{minipage}[b]{0.49\textwidth}
    \includegraphics[width=\textwidth]{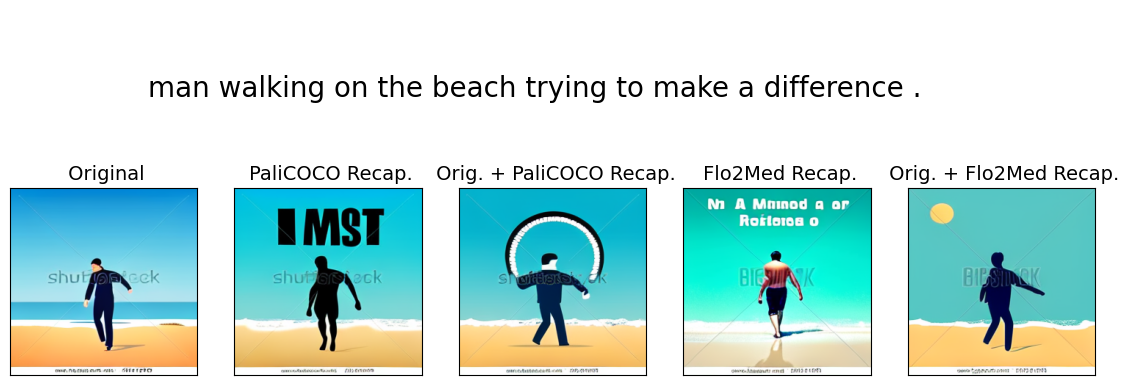}
  \end{minipage}
  \hfill
  \begin{minipage}[b]{0.49\textwidth}
    \includegraphics[width=\textwidth]{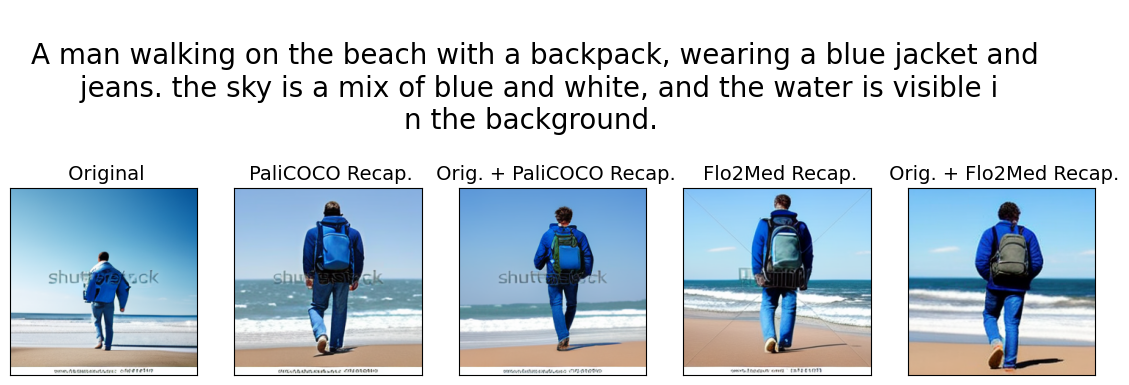}
  \end{minipage}
  \begin{minipage}[b]{0.49\textwidth}
    \includegraphics[width=\textwidth]{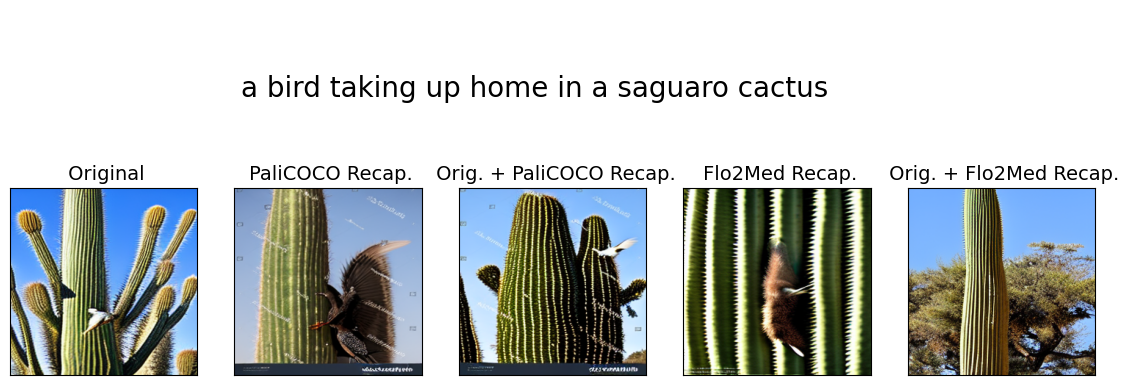}
  \end{minipage}
  \hfill
  \begin{minipage}[b]{0.49\textwidth}
    \includegraphics[width=\textwidth]{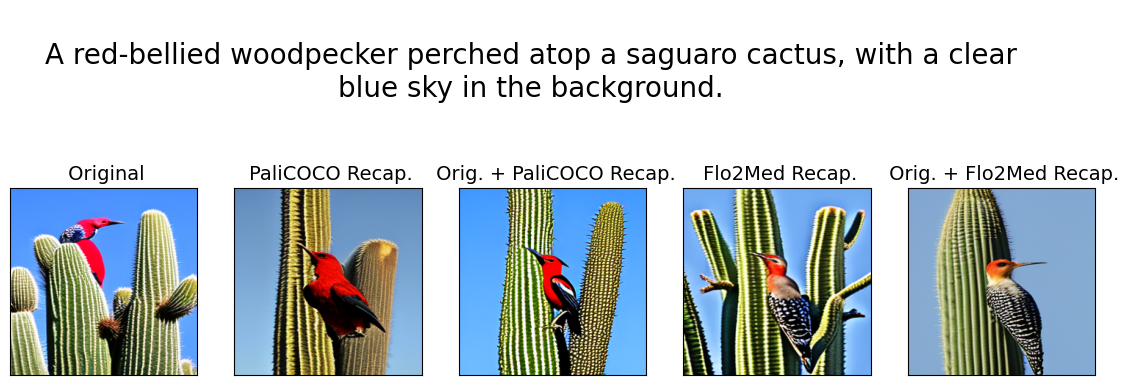}
  \end{minipage}
    \begin{minipage}[b]{0.49\textwidth}
    \includegraphics[width=\textwidth]{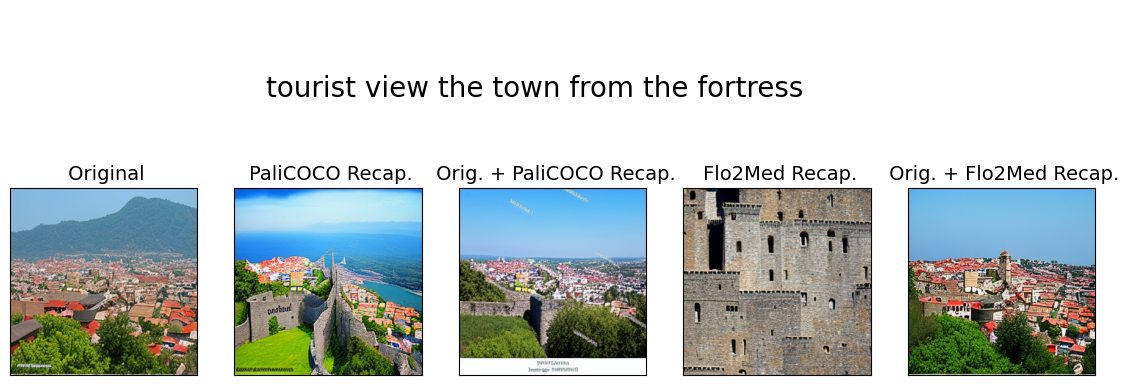}
  \end{minipage}
  \hfill
  \begin{minipage}[b]{0.49\textwidth}
    \includegraphics[width=\textwidth]{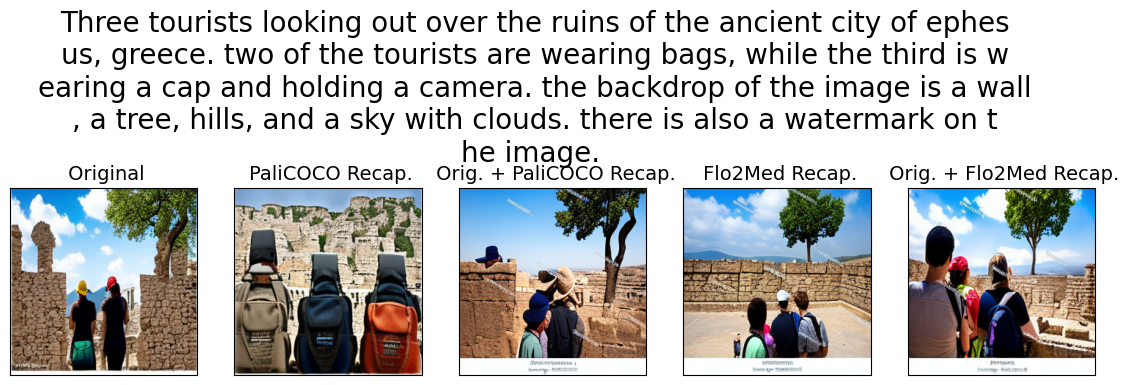}
  \end{minipage}
  \hfill
  \caption{Generations with original CC12M captions \textbf{(LEFT)} and Flo2-Detailed recaptions \textbf{(RIGHT)}.
  Using more descriptive captions when prompting the model can help increase the consistency of the images, even for models trained with coarser captions (e.g. \texttt{CC12M Original} model).}
\end{figure}

Overall, we observe that training models with both original and re-captioned data leads to higher precision and coverage performance across datasets.
Furthermore, we find that re-captioning improves performance for some types of generation tasks more than others, with the stronger performance improvements occurring for the ImageNet and re-captioned CC12M datasets. 
In addition, we find that using only original captions is better than using only recaptioned data when generating images with the original caption distribution.
Excluding original captions can have a larger negative effect on GeoDE and ImageNet.
Furthermore, Florence2 recaptioning alone is worse than PaliCOCO recaptioning.
When inspecting visual examples, we find that re-captioning training data with more dense images and using more dense captions at generation time can help with image consistency, such as increasing the presence of animals or people mentioned in the caption. 
However, using only re-captioned training data for a model can yield undesirable artifacts, like extraneous text or banners, when not prompting with similar prompt styles.

%% file: sections/discussion.tex
\section{Limitations}

While we encourage the use of \repo to benchmark performance, we also discuss useful considerations related to automatic evaluation of text-to-image generative models.

\paragraph{Known limitations with automatic metrics}
Recent works have identified limitations of automatic metrics for text-to-image generative models, such as biases in backbone feature extractors that may prioritize textures~\citep{DBLP:journals/corr/abs-1811-12231}, certain kinds of concepts ~\citep{devries2019doesobjectrecognitionwork}, or stereotypical representations of certain groups~\citep{agarwal2021evaluatingclipcharacterizationbroader}.
In addition, a subset of work is focused on exploring the robustness of consistency metrics~\citep{rossmakes2024,saxon2024evaluates}.
While automatic evaluations of text-to-image generative models are imperfect, they are useful in performing timely and large-scale benchmarking of generative models. 
To enable increasingly robust automatic evaluations, the \repo is designed to enable users to add new metrics as they are developed.

\paragraph{Dataset coverage}
Evaluation datasets are only as effective as their coverage of real world use-cases and people. 
Additional datasets may be added to the \repo to increase evaluation coverage. 
These may build on benchmarks designed specifically for generative modeling to probe skills such as compositional understanding \citep{zhu2023contrastive,huang2023t2i,wu2024conceptmix}, commonsense reasoning \citep{fu2024commonsense} and disambiguation \citep{rassin2022dalle}.
Furthermore, additional datasets that contain per-prompt group information, such as the FACET~\citep{Gustafson_2023_ICCV} dataset depicting people in different activities or professions, may be easily added to unlock new group measurements.

\paragraph{Incorporating human evaluations}
\repo is not intended to replace human evaluations of text-to-image generative models or in-depth qualitative user studies. 
There are several existing frameworks designed for evaluations from humans, where users compare models head-to-head. 
Approaches such as GenAI Arena \citep{jiang2024genai}, K-Sort Arena \citep{li2024k} and Text to Image Arena are dynamic and include humans-in-the-loop. 
GenAI-Bench \citep{li2024genai} provides a large-scale user study using Likert scale to rate generated images along with correlations between humans and automatic metrics. 
Frameworks involving humans approaches are  generally more fine-grained and reliable than automatic metrics, with the tradeoff of being more time consuming and costly.

\section{Conclusion}
In this work, we introduced \repo, a library for evaluating generative models. 
\repo complements existing libraries by including more recent metrics, reproducible visualizations, and Evaluation Exercises that focus benchmarking on specific analysis questions.
In addition, the library has a strong focus on supporting customization with updated or new datasets, metrics, and visualizations to allow for adaptability to future state-of-the-art evaluation methods. 
In this paper, we leveraged \repo to reproduce existing analysis methods focused on Pareto Front-based measurements of trade-offs and disaggregated measurements across geographic groups.
Furthermore, we also introduced two new Evaluation Exercises focused on evaluating the robustness of model rankings and the effect of model interventions on different prompt styles. 
We hope that with the release of  \repo, researchers and practitioners will be able to reliably benchmark text-to-image models with actionable takeaways to guide future areas of development. 

\section{Acknowledgments}
We thank Carolyn Krol for extensive consultation and support throughout this project. 
In addition, we also thank Brian Karrer and Matthew Muckley for their feedback and support throughout the project. 

\newpage

%% file: sections/appendix.tex
\section{Additional details: Datasets}
\label{app:datasets}

We provide additional details about the datasets included in \repo.

\subsection{Real Image Datasets}

We use the term \textit{real image datasets} to define datasets that have real-world images alongside human-written or automatically generated text prompts. 
These datasets can be used for marginal or reference-free metrics.

\paragraph{COCO} MS-COCO~\citep{lin2014microsoft} is an image captioning dataset that is now used as a standard for text-to-image generative models~\citep{betker2023improving}. 
The \repo supports COCO-2014 validation set, which contains approximately 40,000 images with 5 human-written captions per image.
We default to using the first caption as the prompt for each generated image.

\paragraph{ImageNet} ImageNet~\citep{5206848} is a standard classification dataset for computer vision tasks.
It is useful for testing concept generation and utility of generations in downstream learning tasks~\citep{gao2023masked,peebles2023scalable,rombach2022high,brock2018large}.
The \repo contains support for ImageNet for both class-conditional image generation models and text-to-image models (where the prompt corresponds to the image label). 

\paragraph{CC12M} 
The \repo also contains the Conceptual 12M (CC12M) dataset~\citep{changpinyo2021cc12m}, which contains alt-text captions with up to 256 works per images.

\paragraph{GeoDE} 
To provide insights into performance for different geographic regions, the \repo includes support for the GeoDE~\citep{ramaswamy2023geodegeographicallydiverseevaluation} dataset.
GeoDE contains images of objects that were taken by people living in different countries around the world and has been used to evaluate performance gaps across geographic regions ~\citep{hall2024diginevaluatingdisparities,sureddy2024decomposedevaluationsgeographicdisparities}.
We use the balanced version ~\citep{hall2024diginevaluatingdisparities} to ensure a ``fair'' comparison across objects and regions, since different quantities of images correspond to different manifold sizes and can impact measurements. 
This requires omitting objects that have too few samples and extraneous images of objects that had many samples and yields $29160$ images, $180$ images for each of the six regions for $27$ objects. 
In addition, we remove images that contain the tree tag to avoid any spurious correlations with trees in our measurements.

\subsection{Prompt Datasets}

We define \textit{prompt datasets} as datasets that are text-only. These can be used for reference-free metrics only.

\paragraph{PartiPrompts}
The PartiPrompts~\citep{yu2022scalingautoregressivemodelscontentrich} dataset contains $1600$ prompts of varying complexity and theme, spanning topics including ``Imagination,'' ``Fine-grained detail,'' ``Writing and symbols,'' and ``Quantity.''

\paragraph{T2I-Compbench}
T2I-Compbench~\citep{huang2023t2icompbenchcomprehensivebenchmarkopenworld} is a dataset of $3,000$ prompts used to test compositonality skills of text-to-image generative models. 
It includes prompts corresponding to color-, shape-, and texture-binding, spatial- and non-spatial relationships, and complex compositions.

\paragraph{DrawBench}
The DrawBench~\citep{saharia2022photorealistictexttoimagediffusionmodels} dataset contains $200$ prompts used to evaluate text-to-image models in categories including color, counting, (mis)spelling, rarity, positioning, and depictions of text. 

\section{Additional details: Metrics}
\label{app:metrics}

The library provides support for multiple evaluation metrics, including marginal ones, reference-free ones, and grouped ones, which we detail below.

\subsection{Marginal metrics: Image Realism \& Diversity}
Marginal realism and diversity metrics measure model performance by generating images from prompts that correspond to real world images and compare the \emph{marginal} distribution of generated images to the distribution of real images, as opposed to relying on image-to-image metrics.

\paragraph{FID:} 
We include Fréchet Inception Distance (FID)~\citep{heusel2018ganstrainedtimescaleupdate}, a standard  evaluation metric that compares the distribution of generated images to a corresponding set of real images.
Frequently, the generated images are created with prompts that correspond to the real image dataset.
FID is typically interpreted as an indicator of both image quality and diversity.

\paragraph{Precision/Coverage/Recall/Density:} 
To support more detailed insights into trade-offs in model performance, we include precision~\citep{Sajjadi2018_PR} and density~\citep{kynkäänniemi2019improved} metrics to provide an indicator of generated image realism.
Precision quantifies the proportion of generated images that fall within the manifold of real images while density additionally accounts for the \textit{quantity} of real images the generated images are close to.
We include recall~\citep{Sajjadi2018_PR} and coverage~\citep{kynkäänniemi2019improved} as indicators of the diversity of generated images. 
Recall corresponds to the proportion of real images that fall in the manifold of generated images, while coverage measures the proportion of real images whose manifolds contain at least one generated image.
The \repo includes support for the InceptionV3~\citep{szegedy16cvpr} model as the feature extractor for computing these metrics and uses a nearest-neighbors value of $k=3$ to construct manifolds. 

\subsection{Conditional metrics: Image-Text Consistency}
Reference-free metrics evaluate generated images without any grounding on real-world images. 
The \repo includes consistency metrics that compute a score to measure how well a (generated) image $I$ matches a given text prompt $T$. 
These metrics do not require a set of reference images.

\paragraph{CLIPScore:} CLIPScore~\citep{hessel2021clipscore} embeds the generated image and text prompt and uses a CLIP model \citep{Radford2021LearningTV} to measure the cosine similarity between the two embeddings.
As a note, CLIPScore may ``reward'' more stereotypical 
representations~\citep{10.1145/3630106.3658927}.
The \repo includes support for the OpenCLIP~\citep{ilharco_gabriel_2021_5143773} model. 

\paragraph{Davidsonian Scene Graph (DSG):} DSG \citep{cho2023davidsonian} leverages a language model to generate questions based on the text prompt and a visual question answering (VQA) model to answer the generated questions given the generated images. 

\paragraph{VQAScore:} VQAScore~\citep{lin2024evaluating} uses a VQA model to predict the alignment between a text prompt and a generated image. It does so by computing the probability of the answer \texttt{``yes"} to the question, \texttt{``Does this figure show \{text\}?"}